%% file: submit.tex
\documentclass[lettersize,journal]{IEEEtran}
\usepackage{amsmath,amsfonts}
\usepackage{algorithmic}
\usepackage{array}
\usepackage[caption=false,font=normalsize,labelfont=sf,textfont=sf]{subfig}
\usepackage{textcomp}
\usepackage{stfloats}
\usepackage{url}
\usepackage{verbatim}
\usepackage{graphicx}
\usepackage{cite}
\usepackage[caption=false]{subfig}
\usepackage{bbm}
\newcommand{\etal}{\textit{et al.}}
\begin{document}

\title{Active Control Points-based 6DoF Pose Tracking for Industrial Metal Objects}

\author{Chentao Shen, Ding Pan, Mingyu Mei, Zaixing He,
\IEEEmembership{Senior Member, IEEE}, Xinyue Zhao
\thanks{This work was supported in part by the National Natural Science Foundation of China under Grant 52275514 and Grant 52275547, and in part by the Zhejiang Provincial Natural Science Foundation of China under Grant LY21E050021. (Corresponding author: Chentao Shen, email address: shenchentao@zju.edu.cn.)}
\thanks{Chentao Shen is with both Zhejiang University and Westlake Unversity, Ding Pan, Mingyu Mei, Zaxing He, Xinyue Zhao are with the School of Mechanical Engineering, The State Key Lab of Fluid Power and Mechatronic Systems, Zhejiang University, Hangzhou 310058, China.}}


\markboth{arXiv preprint}%
{Shell \MakeLowercase{\textit{et al.}}: A Sample Article Using IEEEtran.cls for IEEE Journals}


\maketitle

\input{sec/0abstract}

\begin{IEEEkeywords}
Pose tracking, reflective texture-less objects, control points, manipulative transformation.
\end{IEEEkeywords}

\input{sec/1intro}
\input{sec/2relatedworks}

\input{sec/3methodology}
\input{sec/4algorithms}
\input{sec/5experiment}
\input{sec/6conclusion}

\bibliographystyle{IEEEtran}
\bibliography{main}


\newpage

\section*{Biography Section}
\begin{IEEEbiography}[{\includegraphics[width=1in,height=1.25in,clip,keepaspectratio]{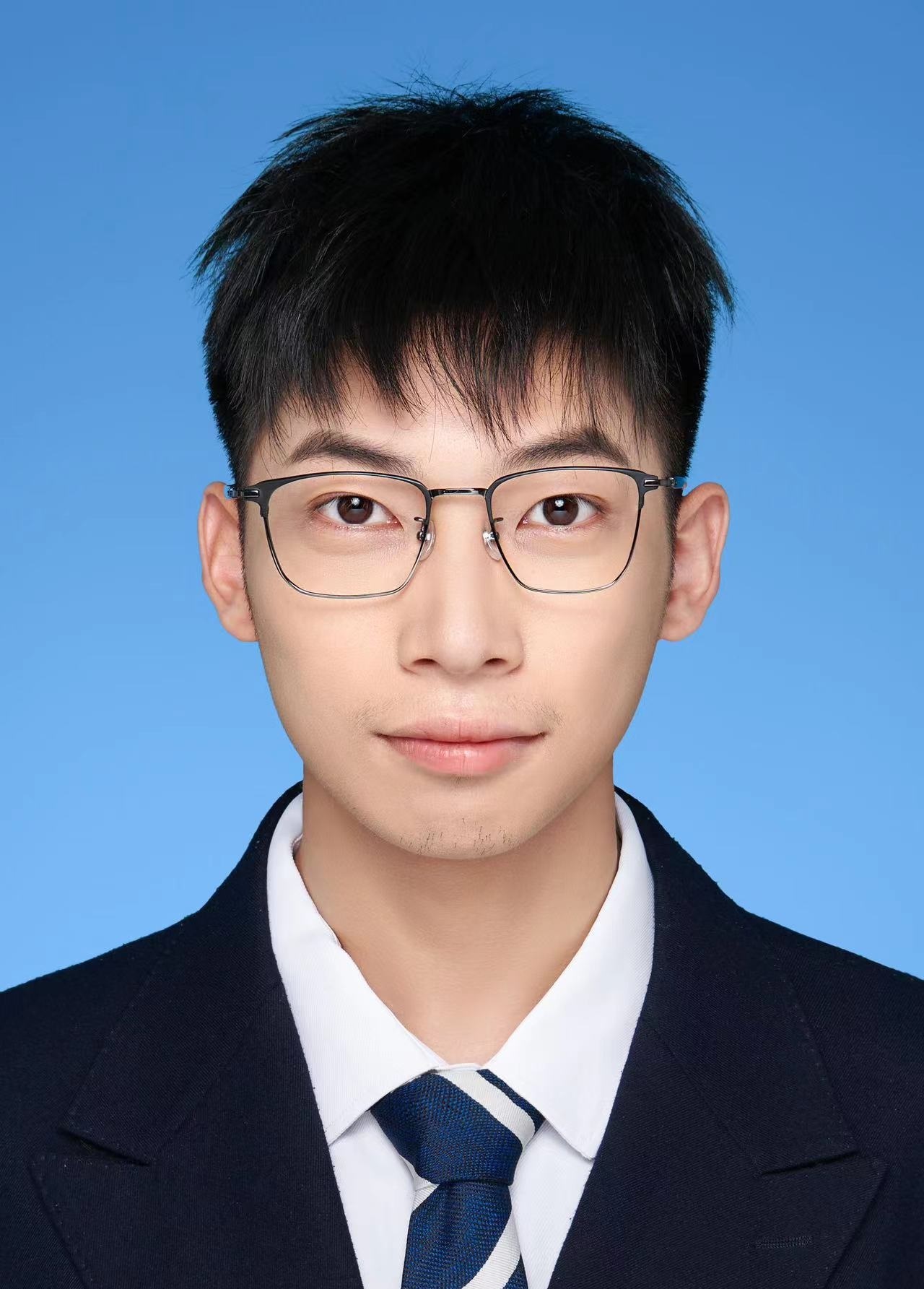}}]{Chentao Shen}
received the B.Sc. and M.Sc. degrees in process Mechanical Engineering from Zhejiang University, China, in 2022 and 2025. He is currently pursuing a doctoral degree in a joint Ph.D. program between Zhejiang University and Westlake University, Hangzhou, China. 
His research interests include robotic vision, computer graphics, and 3D reconstruction.\end{IEEEbiography}

\begin{IEEEbiography}[{\includegraphics[width=1in,height=1.25in,clip,keepaspectratio]{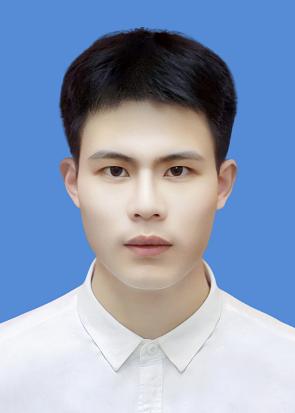}}]{Ding Pan}
received the B.Sc. degree in Measurement, Control Technology and Instruments from Wuhan University of Technology, Wuhan, China, in 2024. He is currently pursuing the M.Sc. degree at Zhejiang University, Hangzhou, China. His research interests include robotic perception and computational imaging.\end{IEEEbiography}
\vspace{-1cm}

\begin{IEEEbiography}[{\includegraphics[width=1in,height=1.25in,clip,keepaspectratio]{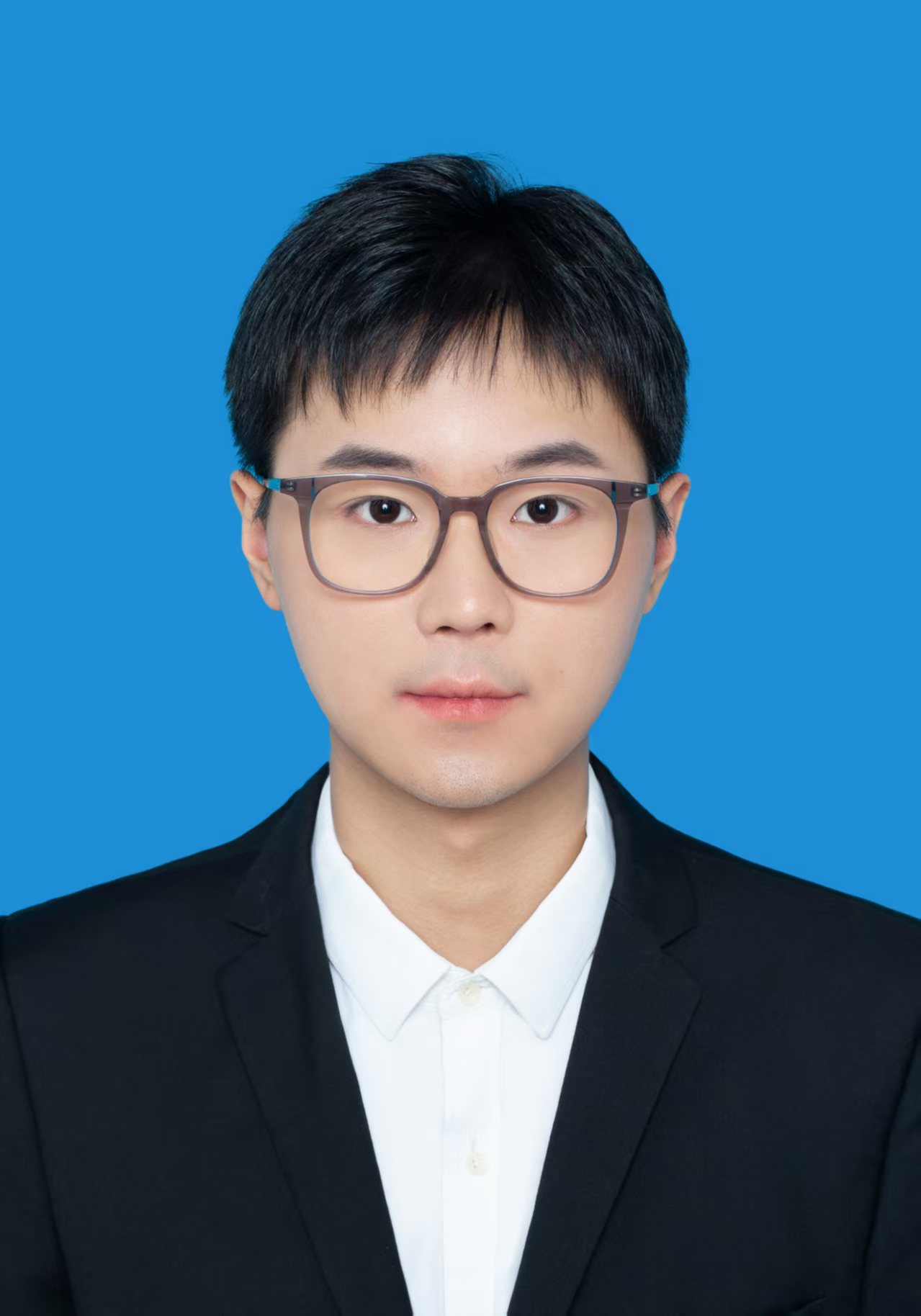}}]{Mingyu Mei}
received the B.Sc. degrees in Electronic Information Engineering from Nanjing University of Post and Telecommunications, China, in 2025. He is currently pursuing a doctoral degree at Zhejiang University, Hangzhou, China.  His research interests include  3D vision, robot learning and vision language action.\end{IEEEbiography}

\vspace{-1cm}
\begin{IEEEbiography}[{\includegraphics[width=1in,height=1.25in,clip,keepaspectratio]{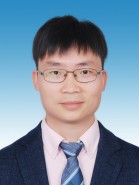}}]{Zaixing He}
(Senior Member, IEEE) received the B.Sc. and M.Sc. degrees in mechanical engineering from Zhejiang University, Hangzhou, China, in 2006 and 2008, respectively, and the Ph.D. degree in information science from the Graduate School of Information Science and Technology, Hokkaido University, Sapporo, Japan, in 2012. 
He is currently an Associate Professor with the School of Mechanical Engineering, Zhejiang University. He has authored/coauthored more than 40 peer-reviewed papers in prestigious journals, such as IEEE TRO, TIE, TII, TIM, IEEE/ASME TMech, Pattern Recognition, and Neurocumputing. His research interests include robotic vision, visual intelligence of manufacturing equipment, and optical-based measurement.
Dr. He served as technical committee members of IEEE Consumer Technology Society and Intelligent Transportation Systems Society, Lead Guest Editor or Guest Editor of several journals including IEEE TCE and Mathematics, Program Chair or TPC of more than 10 international conferences.\end{IEEEbiography}

\vspace{-1cm}
\begin{IEEEbiography}[{\includegraphics[width=1in,height=1.25in,clip,keepaspectratio]{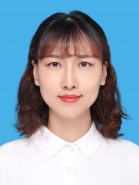}}]{Xinyue Zhao}
received the M.S. degree in mechanical engineering from the Zhejiang University, Hangzhou, China, in 2008, and the Ph.D. degree in information science from the Graduate School of Information Science and Technology, Hokkaido University, Sapporo, Japan, in 2012.
She is currently an Associate Professor with the School of Mechanical Engineering, Zhejiang University. She has authored/coauthored nearly 50 peer-reviewed journal papers. Her research interests include machine vision and image processing.\end{IEEEbiography}

\end{document}

%% file: sec/0abstract.tex
\begin{abstract}
Visual pose tracking is playing an increasingly vital role in industrial contexts in recent years. However, the pose tracking for industrial metal objects remains a challenging task especially in the real world-environments, due to the reflection characteristic of metal objects. To address this issue, we propose a novel 6DoF pose tracking method based on active control points. The method uses image control points to generate edge feature for optimization actively instead of 6DoF pose-based rendering, and serve them as optimization variables. We also introduce an optimal control point regression method to improve robustness. The proposed tracking method performs effectively in both dataset evaluation and real world tasks, providing a viable solution for real-time tracking of industrial metal objects. Our source code is made publicly available at: https://github.com/tomatoma00/ACPTracking.


\end{abstract}

%% file: sec/1intro.tex
\section{Introduction}
\label{sec:intro}

\IEEEPARstart{F}{or} many robotic tasks in industry, such as grasping and assembly, 6DoF (6 Degrees of Freedom) pose estimation is crucial. Methods like BB8~\cite{BB8}, PVNet~\cite{PVNET}, LINEMOD~\cite{LINEMOD}, as well as some methods specifically designed for industrial targets like GFI~\cite{GFI} and ContourPose~\cite{LQZ} have been proposed for pose estimation and shown good performance. For some procedural tasks, such as real-time positioning of a moving target, it is required to ensure real-time performance of pose estimation, which need 6DoF pose tracking. 

In recent years, many researchers have conducted studies on 6DoF pose tracking. Existing approaches are mainly based on one or some of tracking features: keypoints\cite{Y11,Y13}, edges\cite{Q1,Q2,Q3,DAC}, region\cite{PWP,RBOT1,RBGT,SRT}, and deep-learning features\cite{DEEPIM,Y28A}. Considering the characteristics of industrial metal objects, which are primarily composed of geometric elements with a plethora of notable edges, while color information is often not reliable due to reflection characteristics, taking edges as a tracking feature is most suitable. 

However, current edge-based methods exhibit limited robustness during optimization, especially in industrial scenes with complex backgrounds. Specifically, for the majority of optimization process, directly treating the 6DoF pose as independent variables may sometimes lead to following issues.

\begin{figure}[!t]
\centering
\includegraphics[width=3.0in]{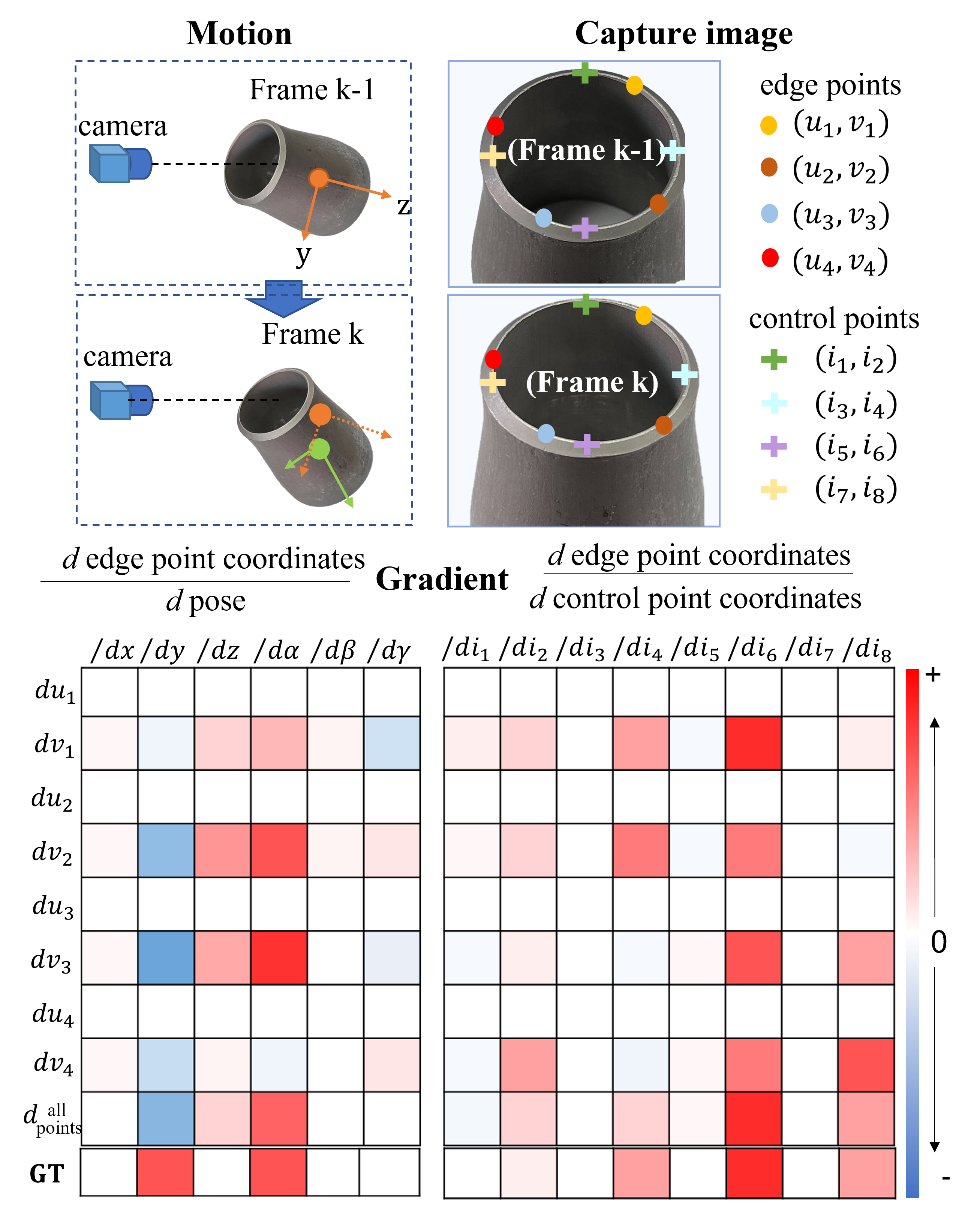}
\caption{The gradient from edge point coordinates to pose and to control points coordinates. The above four images shows the pose change of the object and the corresponding captured image by camera. The below color table shows the value of gradient during the pose change, it shows that the gradient from edge point coordinates to pose is not corresponding to the real pose change, while the control points coordinates corresponds well.}
\label{fig1}
\end{figure}

As the example illustrated in figure \ref{fig1}, the gradient from edge point coordinates relative to 6DoF pose does not accurately reflect the actual pose change (shown in the last two rows of the color table), leading to incorrect iteration directions. Additionally, since pose optimization is based on differences between 2D images, each degree of freedom exhibits varying sensitivities to image changes, complicating the gradient iteration process.

To address these challenges, we propose an noval tracking framework. Drawing inspiration from non-rigid image deformation techniques~\cite{DEFOR}, where selected points are manually moved within an image to induce specific transformations. Regarding the projection of the 3D model under different poses as an image transformation problem, we establish a series of control points, which are utilized to facilitate the generation of image transformations actively. As a problem in image dimension, the gradient reflects actual control points coordinates change more accurately, and the change of the image will be more even as the control points move.

\begin{figure*}[!t]
\centering
\includegraphics[width=5.7in]{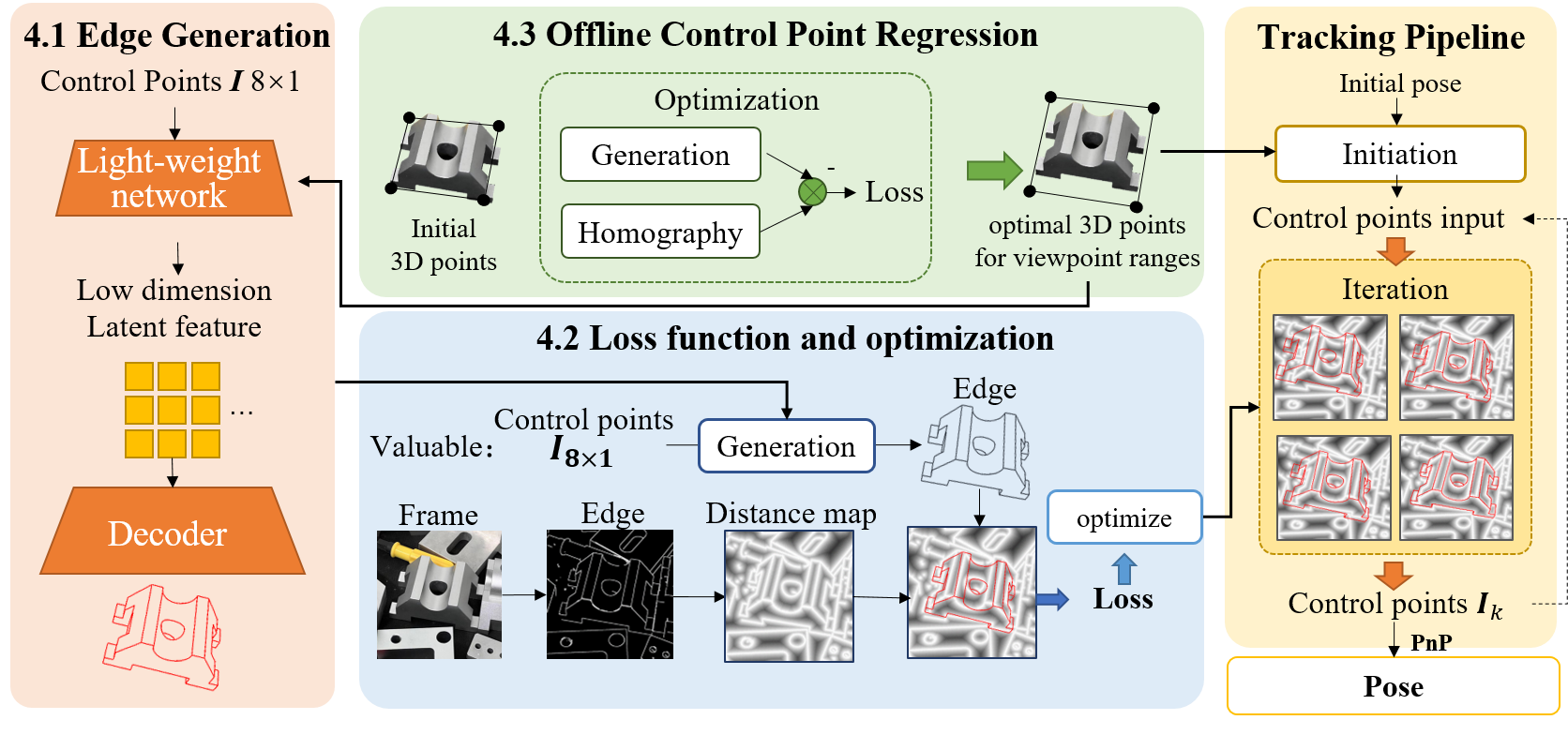}
\caption{overview of the proposed framework. It contains a lightweight model which generates edges, and a loss function for optimization the control points. For each objects, it needs offline optimal control points regression.}
\label{fig4}
\end{figure*}

In general, we propose a novel 6DoF pose tracking algorithm, as shown in figure \ref{fig4}. A lightweight generative model is trained to generate images based on active control points with their 3D information. Then a loss function with control points as independent variables is proposed, which compares the generated edge with captured edge, the function will be optimized in each frame. For 3D information of control points, we regress the optimal value offline.

Our contributions can be summarized as follows:

(1) A novel 6DoF pose tracking framework based on active image control points is proposed, where control points generate the edge feature actively, and be optimized to aligned the edge features with captured edges.

(2) A pose optimization method based on the active control points is proposed, which incorporates a loss function and an edge generation model.

(3) A regression method is proposed to find the optimal control points for an CAD model, aiming to find control points whose coordinate changes produce image modifications that closely approximate an image homography transformation.

The experiments were carried out on both real sequences of industrial metal objects and public dataset, and the results demonstrated that our method has achieved good performance on 6DoF pose tracking of metal objects.

%% file: sec/2relatedworks.tex
\section{Related Works}
\label{sec:rw}

6DoF object tracking is an active research field, there are three mainstream implementations: 

{\bf{Keypoints matching.}} The early keypoint-based methods extract the local image features, followed by matching with template images or keyframes (containing 3D information) to establish correspondences between 2D image points and 3D world points for pose estimation. For instance, Skrypnyk \etal ~\cite{SIFTB} employed SIFT ~\cite{SIFT} for tracking. To enhance the tracking speed, Kim \etal ~\cite{Y11} applied the SIFTGPU, while Vacchetti \etal ~\cite{Y13} used the standard corner detector ~\cite{Y14} for matching. Despite the effectiveness of these methods in certain scenarios, they are limited when dealing with texture-less objects. In recent years, many researchers have turned to deep learning to regress keypoints on texture-less objects. Examples include the works of Hu \etal ~\cite{Y15} and PVNet by Peng \etal ~\cite{PVNET}. However, these approaches typically calculate the pose directly through a single regression of image points, showing no significant advantage in efficiency.

{\bf{ Optimization on region or edges.}} Compared to keypoints, region and edge features are more suitable for texture-less objects. These methods optimize pose to align the generated feature and the captured feature. A classical region-based method is PWP3D~\cite{PWP}, which optimizes the differentiate between the statistical foreground model and the background appearance model. RBOT~\cite{RBOT1,RBOT2}, an extension of PWP3D, using a consistent local color histogram to derive a region-based cost function and employing the Gauss-Newton method for optimization. Recently, Manuel \etal presented RBGT~\cite{RBGT} and SRT3D~\cite{SRT}, introducing multiple viewpoints and a smooth step function, allowing rapid convergence using the Newton method. ICG~\cite{ICG} and ICG+~\cite{ICGP} fuse depth information, enhancing tracking robustness. As for edge-based methods, RAPID~\cite{RAPID} is earliest proposed, it searches significant gradient points near projected edges as control points (different to our approach) and estimated the relative pose through single iteration. Some methods~\cite{Y25,Y26,Y8} improve the RAPID, aiming to reduce the impact of outliers and enhance its robustness. In recent years, Qin \etal ~\cite{Q1,Q2} employed particle filtering during pose initialization to globally find optimal solutions, they also assign the confidence for edges ~\cite{Q3,Q4}, bolstering the method's robustness. DeepAC~\cite{DAC} utilized deep learning to extract more accurate edge feature for tracking.

{\bf{Deep Learning.}} Deep learning based methods regress the deep latent feature in the both captured image and generated image, then compare the feature and calculate the iteration. Examples include DEEPIM~\cite{DEEPIM}, HFF6D~\cite{9792223}, BD-PNP ~\cite{Y27}, and RNNPose~\cite{Y28A,Y28B}. PoseRBPF~\cite{Y29} learns features from images in different rotation and applies particle filtering to solve the global minimal. These methods, compared to directly constructing analytical functions, exhibit greater robustness. Other deep learning features are extracted from 3D point clouds, like~\cite{LIU2024110151,10930556,qin5}, they achieve good results when provided captured 3D points, however, the existing RGB-D camera struggle in scanning industrial mental part.

In industrial environments, where object features are not prominent and high precision is required, keypoints features may not be suitable. As for reflective parts, color information is often not reliable due to highlights and other color variations in reflective areas, making region features less applicable. Therefore, edge features are more suitable for tracking reflective, texture-less parts in industrial settings. There also exists few pose estimation methods for these objects, GFI ~\cite{GFI} use VAE model to create the edge of CAD model and combine the edge of captured image to get the 6DoF pose through optimization, while STB ~\cite{STB} uses geometric features and the correlation of straight contours to represent the part, then matched special location points on the endpoints of the straight contours, to accurately estimate the 6DoF pose. However, these methods exhibit low speed, which is hard to be applied to tracking tasks. 

Therefore, there is still a need for a robust pose optimization method while minimizing computational overhead to address the challenge of 6DoF pose tracking for reflective texture-less objects in industrial settings.





%% file: sec/3methodology.tex
\section{Overview}
\label{sec:method}
This paper proposes a novel pose tracking method designed to provide solutions for industrial objects. An overview of the proposed method is illustrated in figure \ref{fig4}.

As discussed above, 6DoF pose optimization is a common stage in edge feature-based tracking approaches. Typically, this process involves constructing a loss function between the 6DoF pose and the difference between the generated (projected) image and the captured image, then some optimization algorithms like Gauss-Newton, Levenberg-Marquardt~\cite{LM} are employed to find the optimal pose, as (\ref{eq:31}).
\begin {equation}
\label{eq:31}
\mathop{\text{argmin}}_{\boldsymbol{R},\boldsymbol{t}} e(\boldsymbol{R},\boldsymbol{t}), \boldsymbol{R}\in SO(3), \boldsymbol{t}\in \mathbb{R}^{3\times1}
\end {equation}


However, we recognize that directly optimizing the 6DoF pose can sometimes be challenging, especially in obtaining an accurate iteration direction, as the example shown in figure \ref{fig1}. The discrepancy between the 6DoF pose existing in 3D space and the loss function established in 2D space on the imaging plane can lead to inconsistencies in the optimization process, making it easy to obtain the wrong iteration direction. On the other hand, as the viewpoint changes, the impact on projected image of pose parameter changes, it sometimes causing a small gradient while sometimes large. This variability increases the difficulty of optimization.

To address these issues, we treat the projection image of a moving object as a 2D image transformation problem. In the area of image transformation, there exists an approach~\cite{DEFOR} which selects several points in the image and moves them to generate specific image deformation. 

Therefore, we propose a novel framework that optimizes active control points, which generate edges based on their coordinates, to align the generated edges with those in the target image.

\begin{figure}[!t]
\centering
\includegraphics[width=3.5in]{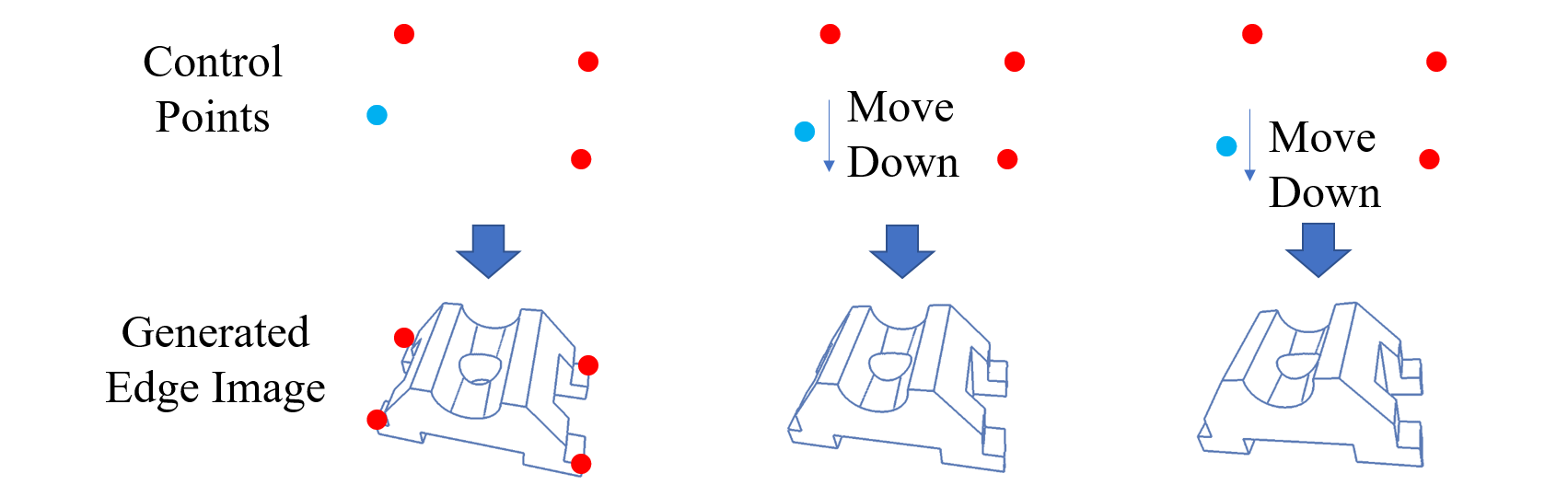}
\caption{An example of manipulative transformation.The image changes by moving of control points (blue) and applying generation model.}
\label{fig3}
\end{figure}

Firstly, we construct a generation model with control points as input, by adjusting the coordinates of control points, achieving transformations of the projection edges, as shown in the figure \ref{fig3}. The detail of edge generation will be introduced in Section \ref{sec:am1}.

Subsequently, we construct a loss function model $ e = E(\boldsymbol{I}) $ as independent variable is the control points $\boldsymbol{I}$, with dimensions of $8$. The function compares the generated edge with the captured edge to calculate the distance between two images. In the tracking process, the loss function is optimized and the optimal control point coordinates are obtained. After completing pose tracking for a frame, we use the output image control points for the initiation of next frame. The pose of each frame could be calculated by solving a pnp (Perspective-n-Point) problem combining with their 3D information. The detail will be introduced in Section \ref{sec:am2}.

To further enhance the robustness of the method, we regress a suitable set of control points, so that the transformations induce by them on the image are closely approximate homographic transformations. In this way, when calculating iteration direction, we can use homographic transformation in pure image to approximate the real transformations caused by the control points, thereby harnessing the advantages of image transformations. This process is integrated into the main tracking procedure, the detail will be introduced in Section \ref{sec:am3}.

Compared to frameworks that adjust other parameters, such as the 6DoF pose of objects, our approach offers a key advantage: the image transformations induced by displacing 2D control points are more independent, it is less likely to be replicated by another movement of control points. As discussed earlier, this independence reduces the risk of the optimization process diverging in erroneous directions over successive iterations.

%% file: sec/4algorithms.tex
\section{Algorithms}
\label{sec:alm}
In this section, the implementation details of the algorithm are introduced.

\subsection{Control Point-based Edge Generation}
\label{sec:am1}

To generate the edge of a CAD model, the most common approach is using rendering tools to render each face and then detect the edges. However, it is inefficient because rendering each face individually is slow, especially when the model contains a large number of faces. 

\begin{figure*}[!t]
\centering
\includegraphics[width=5.7in]{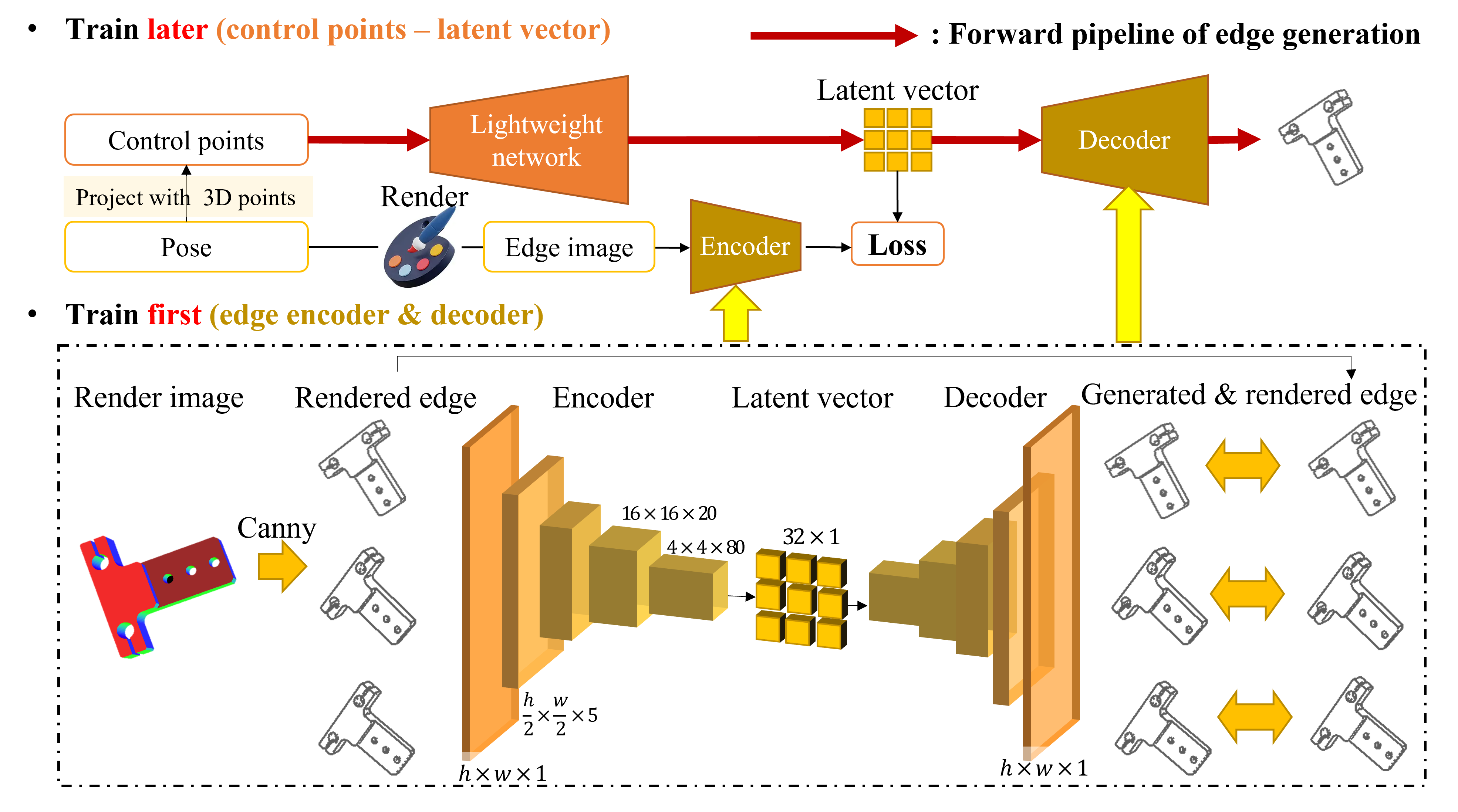}
\caption{The pipeline of our generation model. It contains a lightweight model with control points as input and the latent vector as output, and a decoder with edge image output. We first train the edge image encoder and decoder, then use the decoder for main generation pipeline training.}
\label{fig5}
\end{figure*}

Therefore, we propose a two-stage generation approach, for a given CAD model and four coplanar 3D points. By inputting four image control points, the approach generates an edge image. the detail is shown in figure \ref{fig5}.

First, we encode the edge image into a latent code. As long as the output image dimensions are set, the generation time remains consistent. This method shows better performance when the model has a large number of faces. As for training, we generate a pose set to render about 1000 images, the detect their edges as both the training data and the ground truth. To train the encoder, we combine Binary Cross Entropy loss ($l_{\text{BCE}}$) and Peak Signal-to-Noise Ratio ($l_{\text{PSNR}}$), $\boldsymbol{E}^{\text{GT}}$ represents the ground truth edge image, while $\hat {\boldsymbol{E}}$ indicates the results from encoder and decoder, $\circ$ represents the Hadamard product of two matrics, $N=h\times w$ presents the number of pixels, the weight factors $\lambda$ is set to be 0.5 in practice.

\begin{equation}
    l_{\text{BCE}} = - \frac{1}{N} \text{sum}[{\boldsymbol{E}}^{\text{GT}} \circ \log(\hat {\boldsymbol{E}})+(1-{\boldsymbol{E}}^{\text{GT}})\circ\log(1 -\hat {\boldsymbol{E}})]
\end{equation}
\begin{equation}
    l_{\text{PSNR}} = - \frac{1}{10 \cdot \log_{10}(\frac{1}{N} \|{\boldsymbol{E}}^{\text{GT}}-{\hat E}\|_{2}^2)}
\end{equation}
\begin{equation}
    l_{\text{encoder}} = l_{\text{BCE}} + \lambda l_{\text{PSNR}}
\end{equation}

Subsequently, a lightweight network for regressing the latent code is construct, which enables the process from control points to image generation. We utilize the pose set to project four coplanar 3D points in the image as control points input, while using these pose to render image and encode the latent code as output ground truth, Mean Squared Error (MSE) is used to guide the training.

After training, the forward edge generation process is represented by (\ref{eqtr}), where $\boldsymbol{E}$ is the edge image generated by control points, which is represented as a vector $\boldsymbol{I}$ of dimension 8, where $(I_1,I_2)$ indicates the coordinates of first control point, similarly, $(I_3,I_4), (I_5,I_6), (I_7,I_8)$ indicates the second, third fourth.

\begin{equation}
\label{eqtr}
\boldsymbol{E} = t(\boldsymbol{I}), 
\boldsymbol{I}\in \mathbb{R}^{8}, 
\boldsymbol{E}\in \mathbb{R}^{h\times w}
\end {equation}

\subsection{Loss Function and Pose optimization}
\label{sec:am2}

We construct a loss function $e(\boldsymbol{I})$, to quantify the accuracy of poses generated by different image control points. By optimizing this function, we obtain the optimal control points and 6DoF pose for one frame. 

Firstly, we preprocess the captured image $\boldsymbol{F}\in \mathbb{R}^{h\times w \times 3}$ by applying blur, denoising, and edge extraction (e.g., using the Canny edge detector). This results in a binary edge image $\boldsymbol{E}_{target}\in \mathbb{R}^{h\times w}$ with the same resolution as $\boldsymbol{F}$. Next, we utilize the fast distance transform~\cite{DIST} to build a distance map of the edge image. This map provides the Euclidean distance to the nearest edge for each pixel. We define matrix $\boldsymbol{D}\in \mathbb{R}^{h\times w}$, each element $\boldsymbol{D}_{\boldsymbol{x}}$ represents the distance from pixel $\boldsymbol{x}$ to the nearest edge in $\boldsymbol{E}_{target}$. 


Secondly, for the control points $\boldsymbol{I}$, we use them to generate transformation to obtain the edge image $\boldsymbol{E}$. We map this edge onto the distance map, resulting in the average distance between the projected image and the real image. This value is set as the output, $e$. The smaller the value of $e$, the closer the generated edge is to the captured edge. We denote it as $e(\boldsymbol{I})$. To account for missing edges due to reflection, we introduce a distance threshold $\alpha$, as shown in (\ref{eq8}). If the distance from a edge point to nearest edge is more than $\alpha$, we consider it may encounter the missing edge, setting a lower weight for it. The ratio $\alpha$ is set to be 5\%-10\% of the object diameter for tolerance for occlusion while remaining accuracy, it depends on the complexity of the object, more complex the object is, the smaller $\alpha$ be set. 

\begin {equation}
\label{eq8}
e(\boldsymbol{I}) = \text{sum}{\frac{1}{1-e^{(t(\boldsymbol{I})\circ {\boldsymbol{D} }-\alpha)}}}
\end {equation}

Finally, we optimize the loss function shown in (\ref{eq9}).

\begin {equation}
\label{eq9}
\boldsymbol{I}^{*} = arg min ( {e(\boldsymbol{I})}) 
\end {equation}

The optimization is of 8 dimensions with two redundant degrees of freedom, providing a flexible iterative path to reach the optimal solution.

After optimization, the pose of the object can be easily calculated by solving a PnP problem based on coordinates of control points with their corresponding 3D points.

\subsection{Offline Control Point Regression}
\label{sec:am3}

The selection of control points within different viewpoint ranges is a crucial factor affecting performance. The selection of control points involves determining their corresponding 3D points, which are used to generate edge images. 

\begin{figure*}
\centering
\includegraphics[width=5.4in]{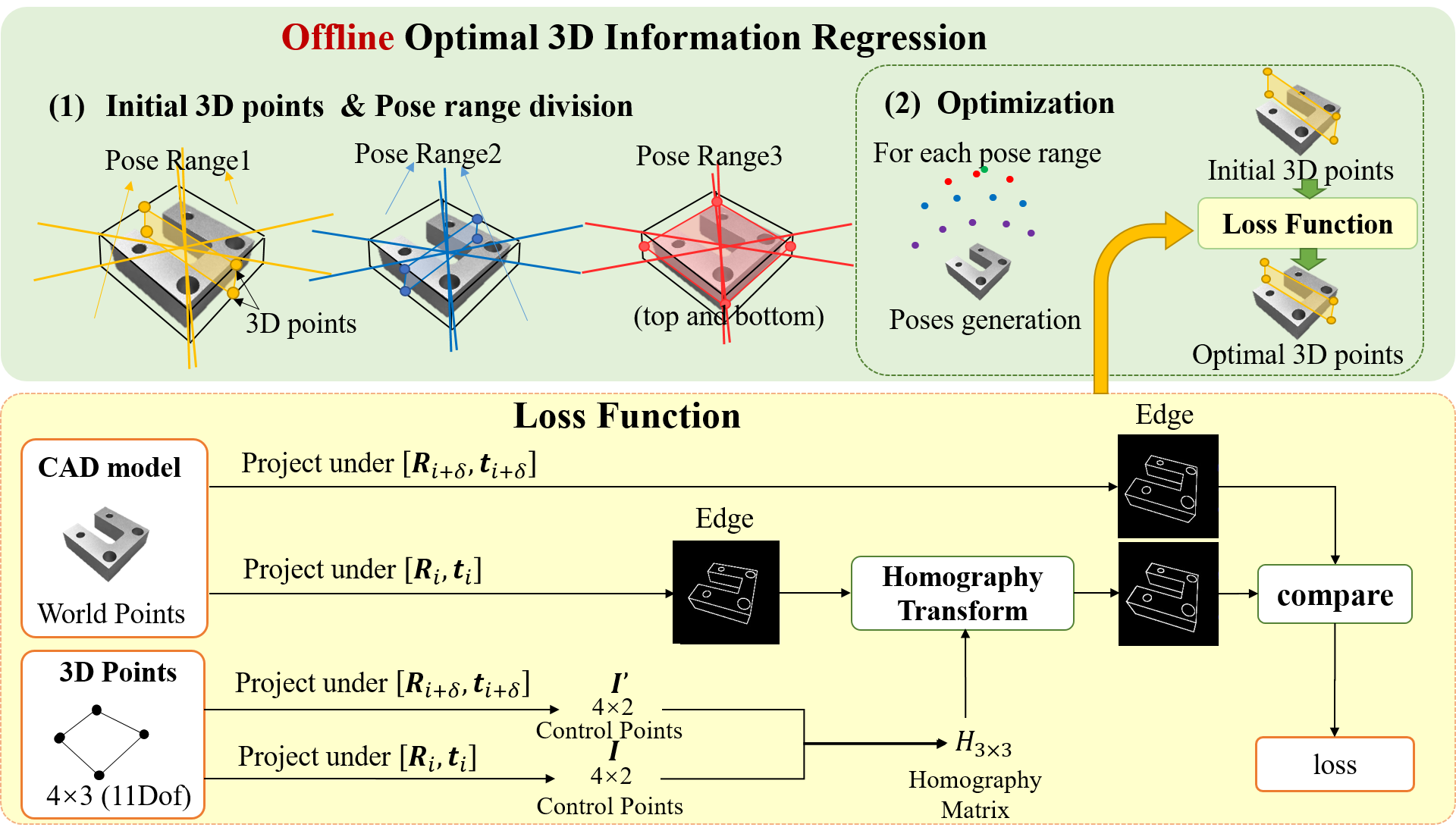}
\caption{Offline control point regression. 3D points sets and pose set are generated, then 3D control points are optimized for each pose range based on loss function. The loss function is aim to compare the effect between homography transformation and real image transformation when under pose change.}
\label{fig6}
\end{figure*}

We consider that a well-selected set of 3D points can make the changes in the edge image caused by the control points smoother, avoiding huge changes in the image due to minor changes in the control points. This can make the optimization process more robust. Therefore, we aim to make the changes in the edge image caused by the control points resemble a homography transformation (a linear mapping). So a coplanar 3D points regression approach is proposed, as illustrated in the figure \ref{fig6}.

Firstly, the pose regions are divided. The minimum bounding box of the object model is computed, ensuring its faces parallel to the coordinate planes. Subsequently, 8 lines are constructed by connecting the geometric center of the object to the vertices of the bounding box, dividing the 3D space into six pose regions.

Secondly, calculate the 3 middle section plane of bounding box, we let the 4 vertices of each section plane as an initial 3D coordinates set, written as $ \boldsymbol{W} = \{ \boldsymbol{P}_{1},\boldsymbol{P}_{2}, \boldsymbol{P}_{3},\boldsymbol{P}_{4}|\boldsymbol{P}_{i}\in\mathbb{R}^{3}\}$. For each pose region, the initial 3D coordinates set is the one not included in this pose region. 

Thirdly, a pose set in the pose region are firstly sampled. We perform uniform sampling at various latitudes and longitudes on a sphere centered at the CAD part. This yields a series of sample points. For the pose of a sample point, its z-axis is directed towards the center of the object, the x-axis is tangent to the latitude in a clockwise direction.

Finally, the 3D coordinates are optimized to better approximate the homography transformation. We construct a loss function, $ e_{3d}(\boldsymbol{W})$, to refine the coordinates of 3D points, treating $\boldsymbol{W}$ as variables, especially, the freedom of the 3D coordinates set is 11 to ensure the coplanarity.

1) For each pose $[\boldsymbol{R}_{i},\boldsymbol{t}_{i}]$ in pose set, we render the projection edge image $\boldsymbol{E}_i^{p}$ of the CAD model by some traditional tools. Then we hypothesize that the object undergoes motion at the pose, then render the object, resulting in the edge image $\boldsymbol{E}_{i+\delta}^{p}$ under the pose after motion, $[\boldsymbol{R}_{i+\delta},\boldsymbol{t}_{i+\delta}]$.  In this paper, the angle and distance change between two pose is set to 0.5°and 5 mm.



2) Subsequently, we calculate the image control points group $\boldsymbol{I}_i$ under pose $[\boldsymbol{R}_{i},\boldsymbol{t}_{i}]$, and $\boldsymbol{I}_{i+\delta}$ under pose $[\boldsymbol{R}_{i+\delta},\boldsymbol{t}_{i+\delta}]$ based on 3D point. We then calculate the homography transformation $\boldsymbol{H}\in\mathbb{R}^{3\times3}$ between control points in $\boldsymbol{I}_i$ and $\boldsymbol{I}_{i+\delta}$ , as shown in (\ref{eq:geth}), while $H_{3,3}=1$. And we apply it to the edge image $\boldsymbol{E}_i^{p}$, resulting in new edge after homography transform $\boldsymbol{E}_{i+\delta}^{h}$, as shown in (\ref{eq:hedge}), where $\mathbb{P}(\cdot)$ indicates Homogeneous coordinate transformation.

\begin {align}
\small{
\begin{bmatrix}
    H_{11}\\H_{12}\\H_{13}\\ \vdots \\ H_{32}
\end{bmatrix} = 
\begin{bmatrix}
I_{i_1}&I_{i_2}&1&0&0&0&-I_{i+\delta_1}I_{i_1}&-I_{i+\delta_1}I_{i_2}\\
0&0&0&I_{i_1}&I_{i_2}&1&-I_{i+\delta_2}I_{i_1}&-I_{i+\delta_2}I_{i_2}\\
I_{i_3}&I_{i_4}&1&0&0&0&-I_{i+\delta_3}I_{i_3}&-I_{i+\delta_3}I_{i_4}\\
&&&&\vdots&&&\\
0&0&0&I_{i_7}&I_{i_8}&1&-I_{i+\delta_8}I_{i_7}&-I_{i+\delta_8}I_{i_8}
\end{bmatrix}^{-1} }
\label{eq:geth}
\end{align}
\begin {equation}
 \boldsymbol{x'} = \mathbb{P}^{-1}(\boldsymbol{H}_{\delta}\cdot \mathbb{P}(\boldsymbol{x})),\boldsymbol{x}\in \boldsymbol{E}_i^{p}, \boldsymbol{x'}\in \boldsymbol{E}_{i+\delta}^{h}
\label{eq:hedge}
\end {equation}

3) Then we compare between the transformed edge points $\boldsymbol{E}_{i+\delta}^{h}$ and the projection edge points $\boldsymbol{E}_{i+\delta}^{p}$ under pose $[\boldsymbol{R}_{i+\delta},\boldsymbol{t}_{i+\delta}]$, calculating the distance between them as the loss of loss function $e_{3d}(\boldsymbol{W})$ based on distance transform as (\ref{eqloss}), here we write the distance map of $\boldsymbol{E}_{i+\delta}^{p}$ as $D_{\boldsymbol{E}_{i+\delta}^{p}}$. A smaller distance indicates that the effect of the control point manipulation transformation is closer to the homography transformation. 

\begin {equation}
\label{eqloss}
e_{3d}(\boldsymbol{W}) =\frac{1}{wh}\|D_{\boldsymbol{E}_{i+\delta}^{p}} \circ\boldsymbol{E}_{i+\delta}^{h}\|_{2}
\end {equation}

By using ADAM to optimize the distance, we obtained the optimal 3D coordinates of control points in a pose region. 

%% file: sec/5experiment.tex
\section{Experiments}
\label{src:exp}
\noindent In this section, we validate the proposed method with a series of experiments. We also compare the proposed method with several state-of-the-art methods in terms of both accuracy and robustness and runtime.

\subsection{Details of the Experiments}
{\bf{Environment.}} Our algorithm is implemented in Python without parallel computing techniques, and runs on an Intel i5-12400 CPU with 16GB RAM. 

{\bf{Dataset.}} The dataset used in experiments includes dataset RT-Less~\cite{RTL} that specially designed for metal objects pose recognition and the public industrial texture-less dataset T-Less~\cite{TLESS}, along with video sets we captured in real-world scene with 1224 × 1024 resolution, the tracking objects captured are sourced from RT-Less~\cite{RTL}. Based on characteristics of the objects in the dataset, we divide them into few groups, objects 1\&2, 3\&4\&6, 15\&17\&34, and 13\&14\&18 in RT-Less form 4 groups of increasing complexity, containing more intricate geometric elements, while objects 6\&7\&9, 20\&23\&26 forming 2 groups based on primary geometry of the objects. The initial pose is set to the ground truth of first frame with a disturbance. 

{\bf{Comparative methods.}} We conducted comparative experiments of the proposed method with various existing approaches suitable for our industrial cases, including classic representatives from each category. These include region-based methods RBGT~\cite{RBGT} and SRT3D~\cite{SRT}, edge-based method GOSTracker~\cite{Q1}, SLET~\cite{Q4}. Additionally, deep learning edge-based method DeepAC~\cite{DAC} was also included. Meanwhile, since there is currently no universal tracking method for reflective texture-less parts at present, we compared the proposed method with the reflective texture-less objects pose detection method, STB~\cite{STB} and GFI~\cite{GFI}.

{\bf{Criteria.}} Regarding the evaluation metrics, we utilize ADD scores~\cite{ADD} with 2\% and 5\% along with run time per frame. ADD score with a\% counts the proportion of the tracking frames whose project error is less than a\% of diameter of the target object. 

\subsection{Accuracy and Run Time Comparisons}
\noindent In this section, we applied our method and comparative methods to two public datasets. Since the image sequences are continuous for pose change, we consider each image as a tracking frame. Table \ref{tab1} and \ref{tab2} present the performance of each method.

\begin{figure}
\centering
\includegraphics[width=2.7in]{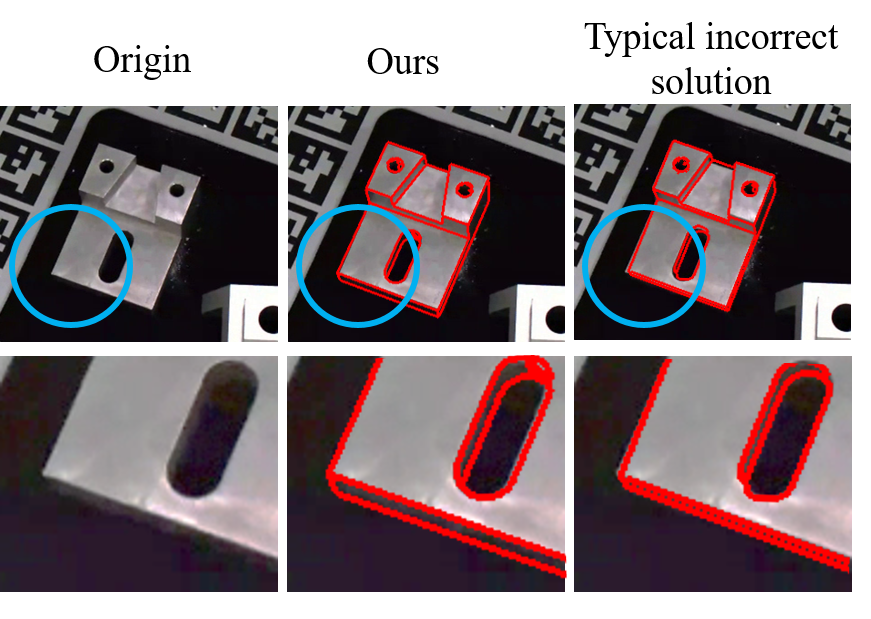}
\caption{A video sequence that the loss of edge has a significant impact on some algorithms. The typical incorrect solution focus more on the region (outermost contour) and fall in to the local solution.}
\label{fig8}
\end{figure}

\begin{table*}
\begin{center}
\caption{The result on ADD score and runtime of RT-Less (metal objects)}
\label{tab1}
\setlength{\tabcolsep}{4.5mm}
\begin{tabular}{ c   c   c   c   c   c   c   c   c   c }
\hline
Criteria & Objects & Ours & RBGT & SRT3D & GOS & SLET & GFI & STB \\
\hline
  & Group1 &$\boldsymbol{0.934}$& 0.722&0.798&0.810&0.826&0.840&0.642\\
ADD & Group2&0.898& 0.774&$\boldsymbol{0.906}$&0.674&0.856&0.766&0.588\\
5\% & Group3 &$\boldsymbol{0.906}$& 0.782& 0.896& 0.646& 0.746& 0.732& 0.440\\
  & Group4&$\boldsymbol{0.688}$&0.354&0.510&0.328&0.532&0.656&0.202\\
\hline
  & Group1 &$\boldsymbol{0.508}$& 0.104&0.198&0.228&0.372&0.368&0.206\\
ADD & Group2&$\boldsymbol{0.488}$&0.316&0.420&0.268&0.384&0.310&0.184\\
2\% & Group3 &$\boldsymbol{0.502}$&0.244&0.366& 0.142& 0.252& 0.368& 0.132\\
  & Group4&0.130&0.044&0.118&0.098&0.116&$\boldsymbol{0.156}$&0.032\\
\hline
runtime&Ave&0.088&0.067&$\boldsymbol{0.042}$&0.099&0.072&0.256&0.040\\
\hline
\end{tabular}
\end{center}
\end{table*}

\begin{table*}
\begin{center}
\caption{The result on ADD score of T-Less (non-metal textureless objects)}
\label{tab2}
\setlength{\tabcolsep}{4.5mm}
\begin{tabular}{ c   c   c   c   c   c   c   c   c   c }
\hline
Criteria & Objects & Ours & RBGT & SRT3D & GOS & SLET & DeepAC \\
\hline
ADD & Group1&0.630&0.558&$\boldsymbol{0.696}$&0.352&0.442&0.670\\
5\% & Group2&0.624&0.518&0.494&0.516&0.502&$\boldsymbol{0.630}$\\
\hline
\end{tabular}
\end{center}
\end{table*}

From the results of RT-Less, it can be observed that the proposed method achieves highest accuracy in the task of tracking. The proposed method shows optimal results in terms of ADD2\%. In the ADD5\% metric, our method and the SRT3D both perform well. However, the region-based method SRT3D loses its effectiveness when dealing with parts group 3 and 4, where contour similarities result in suboptimal performance. In contrast, the proposed method exhibits superior performance in the ADD5\% metric for simple parts (group 1,2) and complex curved parts (group 3,4). GOS and SLET shows good results in parts with simple geometry, they can not work well in complex parts with boss projections. As for other methods, GFI shows better performance, which decreases less in complex parts, however, it costs more time for its training stage. As for dataset of T-Less, though the objects are mainly made of plastics, the proposed methods also shows a good result.

Here we analysis the reason why reflection of metal objects has a significant impact on some algorithms. Figure \ref{fig8} shows a frame of object 3 in RT-Less dataset, at the bottom of the object, due to background lighting and the reflections, it shows dark on the surface. This leads to inaccuracy of the outmost contour and region, making it easily misinterpreted as a situation where the viewing angle is perpendicular to the surface (as seen in the typical incorrect result).

As for run time, SRT3D shows the fastest running speed in our experiments, while our method being approximately on par with that of SLET. Due to the large resolutions of our videos, the tracking process are slower than other datasets, yet the speeds of all tracking methods are fully sufficient within industrial applications.


\subsection{Robustness Comparison}
\noindent In this section, we conducted experiments involving handheld parts. The handheld parts were moved while being captured by a fixed camera. During the moving of handheld parts, various challenging conditions such as shadows, high reflection, and partial reflection were encountered. We compared the proposed method with the best-performing SRT3D method from the previous section.

\begin{figure*}
    \centering
    \subfloat[]{
    \centering
     \includegraphics[width=5.4in]{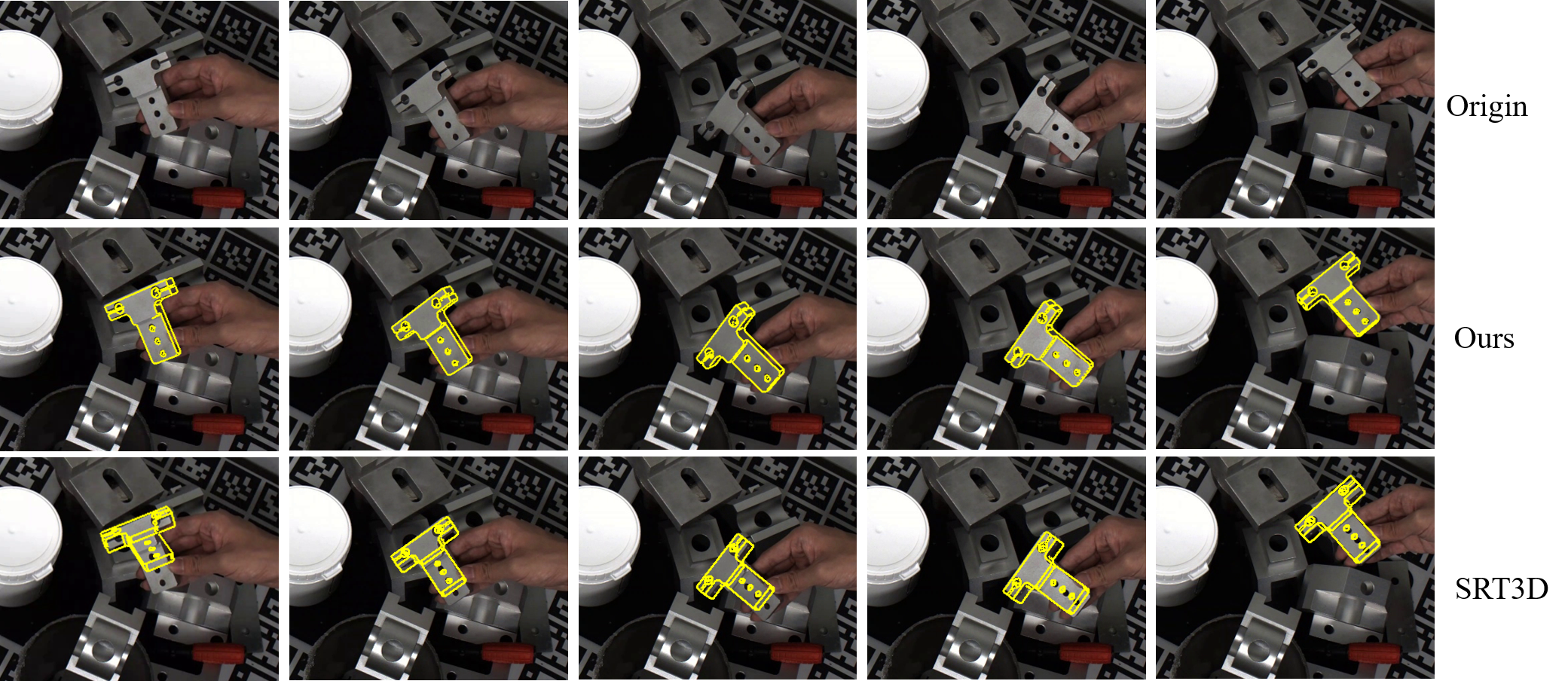}
    \label{fig10}}
  \vfill
  \subfloat[]{
  \centering
     \includegraphics[width=5.4in]{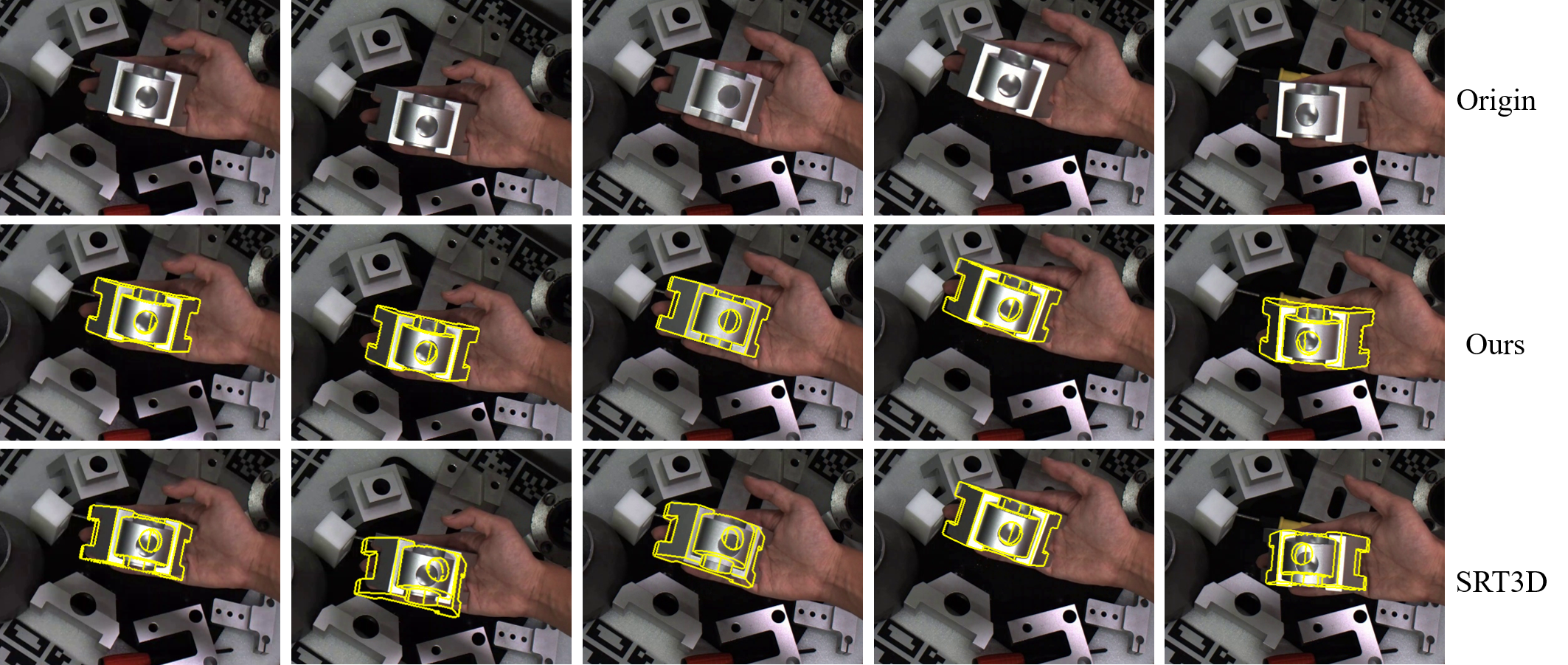}
    \label{fig11}}
    \vfill
  \subfloat[]{
  \centering
     \includegraphics[width=5.4in]{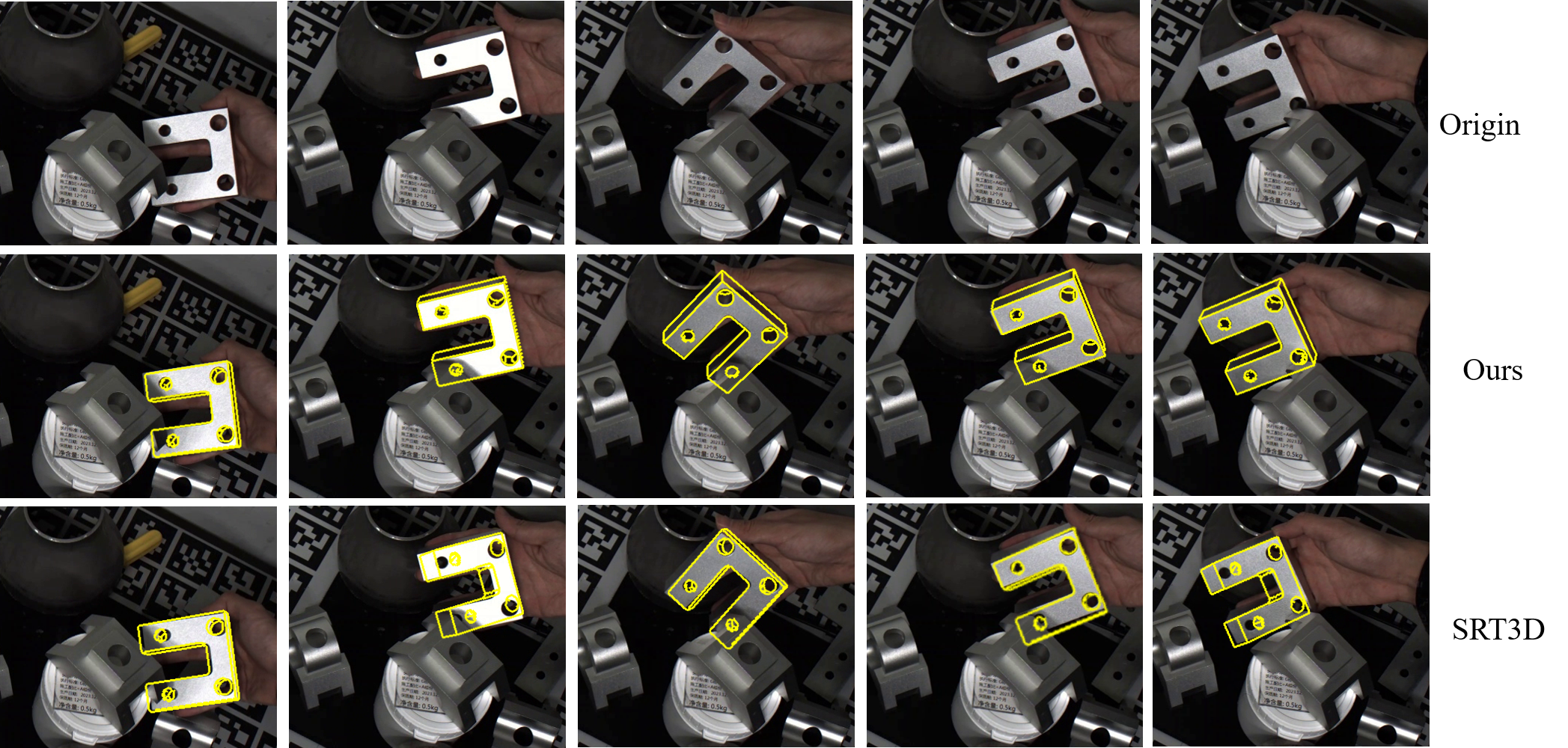}
    \label{fig12}}
  \caption{Experiment result of tracking with handing objects under different conditions. We choose the best-performing method SRT3D for comparison. (a) shows the results under reflection variation. (b) shows the results under pure rotation. (c) shows the results under occlusion condition. }
\end{figure*}




{\bf{Interference of reflection variation.}} We conducted tracking experiments on handheld parts using selected components. Figure \ref{fig10} displays some key frames from the tracking video of object 4. In these frames, due to factors such as reflection, object 4 exhibits different colors at different viewpoints, gradually darkening from the first to the third frame, trending toward the color of the background part. Simultaneously, the side of the part (the surface perpendicular to the reflective surface) presents a dark color due to specular reflection, almost impossible to extract accurate edges, resulting in detection difficulties. In the experiment, SRT3D also demonstrated insufficient robustness, as it couldn't accurately obtain the region of the tracked part, leading to significant errors in some frames.

{\bf{Large Scale Rotation.}} In the experiment, we performed large scale rotations and small scale movements on the object. In the second and third frames of figure \ref{fig11}, the outer contours are extremely similar but the poses are opposite, making traditional methods based on 6DoF pose optimization more prone to incorrect pose. Comparing the proposed method with SRT3D, SRT3D can capture and fit the outer contour well, but its computed pose is incorrect. That's because in these viewpoints, both the incorrect pose and the ground truth have similar outer contours, while the internal color information is not reliable due to reflections.

{\bf{Occlusion.}} This experiment focus on the impact of occlusion. figure \ref{fig12} shows different frames from the video with object 2 for tracking and object 17 for occlusion, each displaying various occlusion scenarios. It is evident that the proposed method exhibits good adaptability to occlusion. SRT3D also shows some adaptability to occlusion, but due to reflection, its performance is not as outstanding.

\subsection{Ablation Study}

We perform ablation experiments on selection of the 3D points of control points described in \ref{sec:am3}, comparing the results with: (1) optimization on 6Dof Pose directly(w/o CP), (2) simply choose the vertices of middle plane of the bounding box as the 3D points(w/o opt). We choose 2 objects (6\&18) with different complexity in RT-Less for this experiments.

\begin{table}[]
    \centering
    \caption{The result on ADD 5\% of ablation study}
    \begin{tabular}{c c c c}
    \hline
        Object&OUR&w/o CP&w/o opt \\
        \hline
         6&0.900&0.660&0.884\\
         18&0.712&0.324&0.646\\
         \hline
    \end{tabular}
    
    \label{tab:as}
\end{table}

The results of the ablation experiments are reported in \ref{tab:as}, where our method outperforms other ablation versions, showing advantages of control points-based optimization and the efficiency of control points regression.





%% file: sec/6conclusion.tex
\section{Conclusion}
\noindent This paper proposes a novel approach for 6DoF tracking of reflective industrial objects based on control points. Image control points are utilized to generate the edge feature, and are served as optimized variables instead of the 6DoF pose. Experimental results demonstrate that the proposed method achieves superior accuracy and robustness in tracking industrial metal objects compared to existing approaches. The proposed method also presents a new perspective on 6DoF tracking by transforming the traditional optimization of 6DoF poses into an optimization for images.

 However, the proposed method is suitable for industrial parts, but showing common results in some objects with curved surface. In the future, we will extend this idea to region-based tracking algorithms, offering a more universally applicable tracking method.


%% file: submit.bbl
\begin{thebibliography}{10}
\providecommand{\url}[1]{#1}
\csname url@samestyle\endcsname
\providecommand{\newblock}{\relax}
\providecommand{\bibinfo}[2]{#2}
\providecommand{\BIBentrySTDinterwordspacing}{\spaceskip=0pt\relax}
\providecommand{\BIBentryALTinterwordstretchfactor}{4}
\providecommand{\BIBentryALTinterwordspacing}{\spaceskip=\fontdimen2\font plus
\BIBentryALTinterwordstretchfactor\fontdimen3\font minus \fontdimen4\font\relax}
\providecommand{\BIBforeignlanguage}[2]{{%
\expandafter\ifx\csname l@#1\endcsname\relax
\typeout{** WARNING: IEEEtran.bst: No hyphenation pattern has been}%
\typeout{** loaded for the language `#1'. Using the pattern for}%
\typeout{** the default language instead.}%
\else
\language=\csname l@#1\endcsname
\fi
#2}}
\providecommand{\BIBdecl}{\relax}
\BIBdecl

\bibitem{BB8}
M.~Rad and V.~Lepetit, ``Bb8: A scalable, accurate, robust to partial occlusion method for predicting the 3d poses of challenging objects without using depth,'' in \emph{ICCV}, 2017, pp. 3848--3856.

\bibitem{PVNET}
S.~Peng, Y.~Liu, Q.~Huang, X.~Zhou, and H.~Bao, ``Pvnet: Pixel-wise voting network for 6dof pose estimation,'' in \emph{CVPR}, 2019, pp. 4556--4565.

\bibitem{LINEMOD}
S.~Hinterstoisser, V.~Lepetit, S.~Ilic, S.~Holzer, G.~Bradski, K.~Konolige, and N.~Navab, ``Model based training, detection and pose estimation of texture-less 3d objects in heavily cluttered scenes,'' in \emph{ACCV}, 2013, pp. 548--562.

\bibitem{GFI}
Z.~He, M.~Wu, X.~Zhao, S.~Zhang, and J.~Tan, ``A generative feature-to-image robotic vision framework for 6d pose measurement of metal parts,'' \emph{IEEE/ASME Transactions on Mechatronics}, vol.~27, no.~5, pp. 3198--3209, 2022.

\bibitem{LQZ}
Z.~He, Q.~Li, X.~Zhao, J.~Wang, H.~Shen, S.~Zhang, and J.~Tan, ``Contourpose: Monocular 6-d pose estimation method for reflective textureless metal parts,'' \emph{IEEE Transactions on Robotics}, vol.~39, no.~5, pp. 4037--4050, 2023.

\bibitem{Y11}
K.~Kim, V.~Lepetit, and W.~Woo, ``Keyframe-based modeling and tracking of multiple 3d objects,'' in \emph{2010 IEEE International Symposium on Mixed and Augmented Reality}, 2010, pp. 193--198.

\bibitem{Y13}
L.~Vacchetti, V.~Lepetit, and P.~Fua, ``Stable real-time 3d tracking using online and offline information,'' \emph{IEEE TPAMI}, vol.~26, no.~10, pp. 1385--1391, 2004.

\bibitem{Q1}
W.~Guofeng, W.~Bin, Z.~Fan, and Q.~Xueying, ``Global optimal searching for textureless 3d object tracking,'' \emph{Vis Comput}, vol.~31, pp. 979--988, 2015.

\bibitem{Q2}
X.~Tian, X.~Lin, F.~Zhong, and X.~Qin, ``Large-displacement 3d object tracking with hybrid non-local optimization,'' in \emph{ECCV}, 2022, pp. 627--643.

\bibitem{Q3}
W.~Guofeng, W.~Bin, Z.~Fan, and Q.~Xueying, ``Robust edge-based 3d object tracking with direction-based pose validation,'' \emph{Multimed Tools Appl}, vol.~78, pp. 12\,307--12\,331, 2019.

\bibitem{DAC}
L.~Wang, S.~Yan, J.~Zhen, Y.~Liu, M.~Zhang, G.~Zhang, and X.~Zhou, ``Deep active contours for real-time 6-dof object tracking,'' in \emph{ICCV}, 2023, pp. 13\,988--13\,998.

\bibitem{PWP}
V.~A. Prisacariu and I.~Reid, ``Pwp3d: Real-time segmentation and tracking of 3d objects,'' \emph{IJCV}, vol.~98, pp. 335--354, 2012.

\bibitem{RBOT1}
H.~Tjaden, U.~Schwanecke, and E.~Schömer, ``Real-time monocular pose estimation of 3d objects using temporally consistent local color histograms,'' in \emph{ICCV}, 2017, pp. 124--132.

\bibitem{RBGT}
M.~Stoiber, M.~Pfanne, R.~Strobl, Klaus H.and~Triebel, and A.~Albu-Sch{\"a}ffer, ``A sparse gaussian approach to region-based 6dof object tracking,'' in \emph{ACCV}, 2021, pp. 666--682.

\bibitem{SRT}
------, ``Srt3d: A sparse region-based 3d object tracking approach for the real world,'' \emph{IJCV}, vol. 130, pp. 1008--1030, 2022.

\bibitem{DEEPIM}
Y.~Li, G.~Wang, X.~Ji, Y.~Xiang, and D.~Fox, ``Deepim: Deep iterative matching for 6d pose estimation,'' in \emph{ECCV}, 2018, pp. 695--711.

\bibitem{Y28A}
Y.~Xu, K.-Y. Lin, G.~Zhang, X.~Wang, and H.~Li, ``Rnnpose: Recurrent 6-dof object pose refinement with robust correspondence field estimation and pose optimization,'' in \emph{CVPR}, 2022, pp. 14\,860--14\,870.

\bibitem{DEFOR}
F.~Bookstein, ``Principal warps: thin-plate splines and the decomposition of deformations,'' \emph{IEEE TPAMI}, vol.~11, no.~6, pp. 567--585, 1989.

\bibitem{SIFTB}
I.~Skrypnyk and D.~Lowe, ``Scene modelling, recognition and tracking with invariant image features,'' in \emph{Third IEEE and ACM International Symposium on Mixed and Augmented Reality}, 2004, pp. 110--119.

\bibitem{SIFT}
D.~Lowe, ``Distinctive image features from scale-invariant keypoints,'' \emph{IJCV}, vol.~60, pp. 91--110, 2004.

\bibitem{Y14}
A.~W. Fitzgibbon and A.~Zisserman, ``Automatic camera recovery for closed or open image sequences,'' in \emph{ECCV}, 1998, pp. 311--326.

\bibitem{Y15}
Y.~Hu, P.~Fua, W.~Wang, and M.~Salzmann, ``Single-stage 6d object pose estimation,'' in \emph{CVPR}, 2020, pp. 2927--2936.

\bibitem{RBOT2}
H.~Tjaden, U.~Schwanecke, E.~Schömer, and D.~Cremers, ``A region-based gauss-newton approach to real-time monocular multiple object tracking,'' \emph{IEEE TPAMI}, vol.~41, no.~8, pp. 1797--1812, 2019.

\bibitem{ICG}
M.~Stoiber, M.~Sundermeyer, and R.~Triebel, ``Iterative corresponding geometry: Fusing region and depth for highly efficient 3d tracking of textureless objects,'' in \emph{CVPR}, 2022, pp. 6845--6855.

\bibitem{ICGP}
M.~Stoiber, M.~Elsayed, A.~E. Reichert, F.~Steidle, D.~Lee, and R.~Triebel, ``Fusing visual appearance and geometry for multi-modality 6dof object tracking,'' in \emph{2023 IEEE/RSJ International Conference on Intelligent Robots and Systems (IROS)}, 2023, pp. 1170--1177.

\bibitem{RAPID}
C.~Harris and C.~Stennett, ``Rapid - a video rate object tracker,'' in \emph{Proceedings of the British Machine Vision Conference}, 1990, pp. 15.1--15.6.

\bibitem{Y25}
G.~Simon and M.-O. Berger, ``A two-stage robust statistical method for temporal registration from features of various type,'' in \emph{ICCV}, 1998, pp. 261--266.

\bibitem{Y26}
T.~Drummond and R.~Cipolla, ``Real-time visual tracking of complex structures,'' \emph{IEEE TPAMI}, vol.~24, no.~7, pp. 932--946, 2002.

\bibitem{Y8}
H.~Wuest, F.~Vial, and D.~Stricker, ``Adaptive line tracking with multiple hypotheses for augmented reality,'' in \emph{Fourth IEEE and ACM International Symposium on Mixed and Augmented Reality (ISMAR'05)}, 2005, pp. 62--69.

\bibitem{Q4}
W.~Bin, Z.~Fan, S.~Yuqing, and Q.~Xueying, ``An occlusion-aware edge-based method for monocular 3d object tracking using edge confidence,'' \emph{Comput. Graph. Forum.}, vol.~39, no.~7, pp. 399--409, 2020.

\bibitem{9792223}
J.~Liu, W.~Sun, C.~Liu, X.~Zhang, S.~Fan, and W.~Wu, ``Hff6d: Hierarchical feature fusion network for robust 6d object pose tracking,'' \emph{IEEE Transactions on Circuits and Systems for Video Technology}, vol.~32, no.~11, pp. 7719--7731, 2022.

\bibitem{Y27}
L.~Lipson, Z.~Teed, A.~Goyal, and J.~Deng, ``Coupled iterative refinement for 6d multi-object pose estimation,'' in \emph{CVPR}, 2022, pp. 6718--6727.

\bibitem{Y28B}
Y.~Xu, K.-Y. Lin, G.~Zhang, X.~Wang, and H.~Li, ``Rnnpose: 6-dof object pose estimation via recurrent correspondence field estimation and pose optimization,'' \emph{IEEE TPAMI}, vol.~46, no.~7, pp. 4669--4683, 2024.

\bibitem{Y29}
X.~Deng, A.~Mousavian, Y.~Xiang, F.~Xia, T.~Bretl, and D.~Fox, ``Poserbpf: A rao–blackwellized particle filter for 6-d object pose tracking,'' \emph{IEEE Transactions on Robotics}, vol.~37, no.~5, pp. 1328--1342, 2021.

\bibitem{LIU2024110151}
\BIBentryALTinterwordspacing
Z.~Liu, Q.~Wang, D.~Liu, and J.~Tan, ``Pa-pose: Partial point cloud fusion based on reliable alignment for 6d pose tracking,'' \emph{Pattern Recognition}, vol. 148, p. 110151, 2024. [Online]. Available: \url{https://www.sciencedirect.com/science/article/pii/S0031320323008488}
\BIBentrySTDinterwordspacing

\bibitem{10930556}
Y.~Yang, W.~Li, Y.~Yao, B.~Zhou, and B.~Fan, ``3d single object tracking with cross-modal fusion conflict elimination,'' \emph{IEEE Robotics and Automation Letters}, vol.~10, no.~5, pp. 4826--4833, 2025.

\bibitem{qin5}
X.~Cao, J.~Li, P.~Zhao, J.~Li, and X.~Qin, ``Corr-track: Category-level 6d pose tracking with soft-correspondence matrix estimation,'' \emph{IEEE Transactions on Visualization and Computer Graphics}, vol.~30, no.~5, pp. 2173--2183, 2024.

\bibitem{STB}
Z.~He, Z.~Jiang, X.~Zhao, S.~Zhang, and C.~Wu, ``Sparse template-based 6-d pose estimation of metal parts using a monocular camera,'' \emph{IEEE Transactions on Industrial Electronics}, vol.~67, no.~1, pp. 390--401, 2020.

\bibitem{LM}
K.~Levenberg, ``A method for the solution of certain non-linear problems in least squares,'' \emph{Quarterly of Applied Mathematics}, vol.~2, pp. 164--168, 1944.

\bibitem{DIST}
F.~PF and H.~DP, ``Distance transforms of sampled functions,'' \emph{Theory of Computing}, vol.~8, no.~19, p. 415–428, 2004.

\bibitem{RTL}
Z.~Xinyue, L.~Quanzhi, C.~Yue, W.~Quanyou, H.~Zaixing, and L.~Dong, ``Rt-less: a multi-scene rgb dataset for 6d pose estimation of reflective texture-less objects,'' \emph{Vis Comput}, vol.~40, p. 5187–5200, 2024.

\bibitem{TLESS}
T.~Hodan, P.~Haluza, S.~Obdržálek, J.~Matas, M.~Lourakis, and X.~Zabulis, ``T-less: An rgb-d dataset for 6d pose estimation of texture-less objects,'' in \emph{2017 IEEE Winter Conference on Applications of Computer Vision (WACV)}, 2017, pp. 880--888.

\bibitem{ADD}
X.~Y., S.~T., N.~V., and F.~D., ``Posecnn: A convolutional neural network for 6d object pose estimation in cluttered scenes,'' in \emph{Robotics: Science and Systems (RSS)}, 2018.

\end{thebibliography}
